\documentclass[10pt,twocolumn,letterpaper]{article}

\usepackage{iccv}
\usepackage{times}
\usepackage{epsfig}
\usepackage{graphicx}
\usepackage{amsmath}
\usepackage{amssymb}
\usepackage[breaklinks=true,bookmarks=false, colorlinks, citecolor=green]{hyperref}

\usepackage[breaklinks=true,bookmarks=false]{hyperref}

\iccvfinalcopy 

\def\iccvPaperID{1406} 

\begin{document}

\title{TALL: Temporal Activity Localization via Language Query}

\author{Jiyang Gao$^1$ \qquad Chen Sun$^2$ \qquad Zhenheng Yang$^{1}$ \qquad Ram Nevatia$^1$ \\
$^1$University of Southern California \qquad $^2$Google Research \\
{\tt\small \{jiyangga, zhenheny, nevatia\}@usc.edu,\quad chensun@google.com} 
}


\maketitle

\begin{abstract}
This paper focuses on temporal localization of actions in untrimmed videos. Existing methods typically train classifiers for a pre-defined list of actions and apply them in a sliding window fashion. However, activities in the wild consist of a wide combination of actors, actions and objects; it is difficult to design a proper activity list that meets users' needs. We propose to localize activities by natural language queries. Temporal Activity Localization via Language (TALL) is challenging as it requires: (1) suitable design of text and video representations to allow cross-modal matching of actions and language queries; (2) ability to locate actions accurately given features from sliding windows of limited granularity. We propose a novel Cross-modal Temporal Regression Localizer (CTRL) to jointly model text query and video clips, output alignment scores and action boundary regression results for candidate clips. For evaluation, we adopt TaCoS dataset, and build a new dataset for this task on top of Charades by adding sentence temporal annotations, called Charades-STA. We also build complex sentence queries in Charades-STA for test. Experimental results show that CTRL outperforms previous methods significantly on both datasets.
\end{abstract}

\section{Introduction}



Activities in the wild consist of a diverse combination of actors, actions and objects over various periods of time. Earlier work focused on classification of video clips that contained a single activity, i.e. where the videos were trimmed. Recently, there has also been significant work in \emph{localizing} activities in longer, untrimmed videos \cite{Singh_2016_CVPR, Ma_2016_CVPR}. 
One major limitation of existing action localization methods is that they are restricted to pre-defined list of actions. Although the lists of activities can be relatively large \cite{caba2015activitynet}, they still face difficulty in covering complex activity questions, for example,  ``A person runs to the window and then look out." , as shown in Figure 1. Hence, it is desirable to use natural language queries to localize activities. Use of natural language not only allows for an open set of activities but also natural specification of additional constraints, including objects and their properties as well as relations between the involved entities. We propose the task of Temporal Activity Localization via Language (TALL): given a temporally untrimmed video and a natural language query, the goal is to determine the start and end times for the described activity inside the video.

For traditional temporal action localization, most current approaches  \cite{Singh_2016_CVPR, Ma_2016_CVPR, Shou_2016_CVPR, Yeung_2016_CVPR, Yuan_2016_CVPR} apply activity classifiers trained with optical flow-based methods \cite{wang2011action, simonyan2014two} or Convolutional Neural Networks (CNNs) \cite{simonyan2014very, tran2015learning} in a sliding window fashion. A direct extension to support natural language query is to map the queries into a discrete label space. However, it is non-trivial to design a label space which has enough coverage for such activities without losing useful details in users' queries.

\begin{figure}
\centering
\includegraphics[scale=0.5]{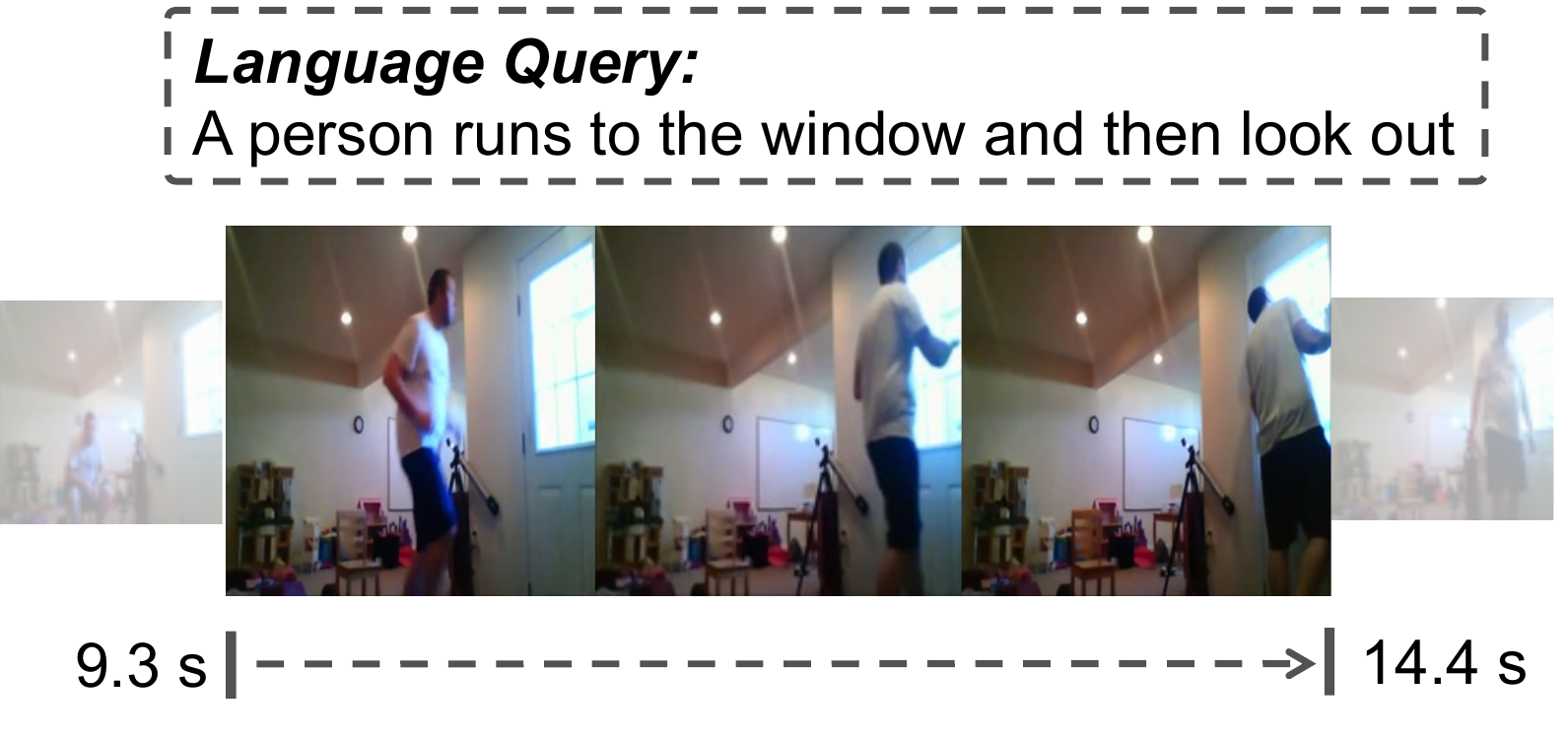}
\caption{ Temporal activity localization via language query in an untrimmed video.}
\end{figure}

To go beyond discrete activity labels, one possible solution is to embed visual features and sentence features into a common space~\cite{Karpathy_2014_CVPR, kiros2015skip, mao2014deep}. However, for temporal localization of activities, it is unclear what a proper visual model to extract visual features for retrieval is, and how to achieve high precision of predicted start/end time. Although one could densely sample sliding windows at different scales, doing so is not only computationally expensive but also makes the alignment task more challenging, as the search space increases. An alternative to dense sampling is to adjust the temporal boundaries of proposals by learning regression parameters; such an approach has been successful for object localization, as in \cite{ren2015faster}. However, temporal regression has not been attempted in the past work and is more difficult as the activities are characterized by a spatio-temporal volume, which may lead to more background noise. 

These challenges motivate us to propose a novel Cross-modal Temporal Regression Localizer (CTRL) model to jointly model text query, video clip candidates and their temporal context information to solve the TALL task. CTRL generates alignment scores along with location regression results for candidate clips. It utilizes a CNN model to extract visual features of the clips and a Long Short-term Memory (LSTM) network to extract sentence embeddings. A cross-modal processing module is designed to jointly model the text and visual features, which calculates element-wise addition, multiplication and direct concatenation. Finally, multilayer networks are trained for visual-semantic alignment and clip location regression. We design the non-parameterized and parameterized location offsets for temporal coordinate regression. In parameterized setting, the length and the central coordinate of the clip is first parameterized by the ground truth length and coordinate. In non-parameterized setting, the start and end coordinates are used directly. We show that the non-parameterized one works better, unlike the case for object boundary regression. 

To facilitate research of TALL, we also generate sentence temporal annotations for Charades \cite{sigurdsson2016hollywood} dataset. We name it  Charades-STA.
We evaluate our methods on TACoS and Charades-STA datasets by the metric of ``R@n, IoU=m", which represents the percentage of at least one of the top-n results ( start and end pairs ) having IoU with the ground truth larger than m.  Experimental results demonstrate the effectiveness of our proposed CTRL framework. 

In summary, our contributions are two-fold:

(1) We propose a novel problem formulation of Temporal Activity Localization via natural Language (TALL) query. 

(2) We introduce an effective Cross-modal Temporal Regression Localizer (CTRL) which estimates alignment scores and temporal action boundary by jointly modeling language query and video clips.\footnote{Source codes are available in {\color{magenta}{https://github.com/jiyanggao/TALL}} .}


\section{Related Work}
\textbf{Action classification and temporal localization.} There have been tremendous explorations about action classification in videos using deep convolutional neural networks (ConvNets). Representative methods include two-stream ConvNets, C3D (3D ConvNets) and 2D ConvNets with temporal LSTM or mean pooling. Specifically, Simonyan and Zisserman \cite{simonyan2014two}  modeled the appearance and motion information in two 
separate ConvNets and combined the scores by late fusion. Tran \emph{et al.} \cite{tran2015learning} used  3D convolutional filters to capture motion information in neighboring frames. \cite{yue2015beyond} \cite{Karpathy_2014_CVPR} proposed to use 2D ConvNets to extract deep features for one frame and use temporal mean pooling or LSTM to model temporal information.

For temporal action localization task, Shou \emph{et al.} \cite{Shou_2016_CVPR} trained C3D \cite{tran2015learning} with localization loss and achieved state-of-the-art performance on THUMOS 14. Ma \emph{et al.} \cite{Ma_2016_CVPR} used a temporal LSTM to generate frame-wise prediction scores and then merged the detection intervals based on the predictions. Singh \emph{et al.} \cite{Singh_2016_CVPR} extended two-stream \cite{simonyan2014two} framework with person detection and bi-directional LSTMs and achieved state-of-the-art performance on MPII-Cooking dataset \cite{rohrbach2012database}. Gao \emph{et al.} \cite{gao2017turn} proposed to use temporal coordinate regression to refine action boundary for temporal localization. 

\textbf{Sentence-based image/video retrieval.} Given a set of candidate videos/images and a sentence query, this task requires retrieving the videos/images that match the query. Karpathy \emph{et al.} \cite{karpathy2015deep} proposed Deep Visual-Semantic Alignment (DVSA) model. DVSA used bidirectional LSTMs to encode sentence embeddings, and R-CNN object detectors \cite{Girshick_2014_CVPR} to extract features from object proposals. Skip-thought \cite{kiros2015skip} learned a Sent2Vec model by applying skip-gram \cite{mikolov2013distributed} on sentence level and achieved top performance in sentence-based image retrieval task. Sun \emph{et al.} \cite{sun2015automatic} proposed to discover visual concepts from image-sentence pairs and apply the concept detectors for image retrieval.  Gao \emph{et al.} \cite{gao2016acd} proposed to learn verb-object pairs as action concepts from image-sentence pairs. Hu \emph{et al.} \cite{hu2015natural} and Mao \emph{et al.} \cite{Mao_2016_CVPR} formulated the problem of natural language object retrieval. 

As for video retrieval, Lin \emph{et al.} \cite{Lin_2014_CVPR} parsed the sentence descriptions into a semantic graph, which are then matched to visual concepts in the videos by generalized bipartite matching.  Bojanowski \emph{et al.} \cite{bojanowski2015weakly} tackled the problem of video-text alignment: given a video and a set of sentences with temporal ordering, assigning a temporal interval for each sentence. In our settings, only one sentence query is input to the system and temporal ordering is not used.

\textbf{Object detection.} Our work is partly inspired by the success of recent object detection approaches. R-CNN \cite{Girshick_2014_CVPR} consists of selective search, CNN feature extraction, SVM classification and bounding box regression.  Fast-RCNN \cite{girshick2015fast} designs RoI pooling layer and the model could be trained by end-to-end framework. One of the key element shared in those successful object detection frameworks \cite{Redmon_2016_CVPR, ren2015faster, girshick2015fast} is the bounding box regression layer. We show that, unlike object boundary regression using parameterized offsets, non-parameterized offsets work better for action boundary regression.
\section{Methods}
\begin{figure*}
\centering
\includegraphics[scale=0.54]{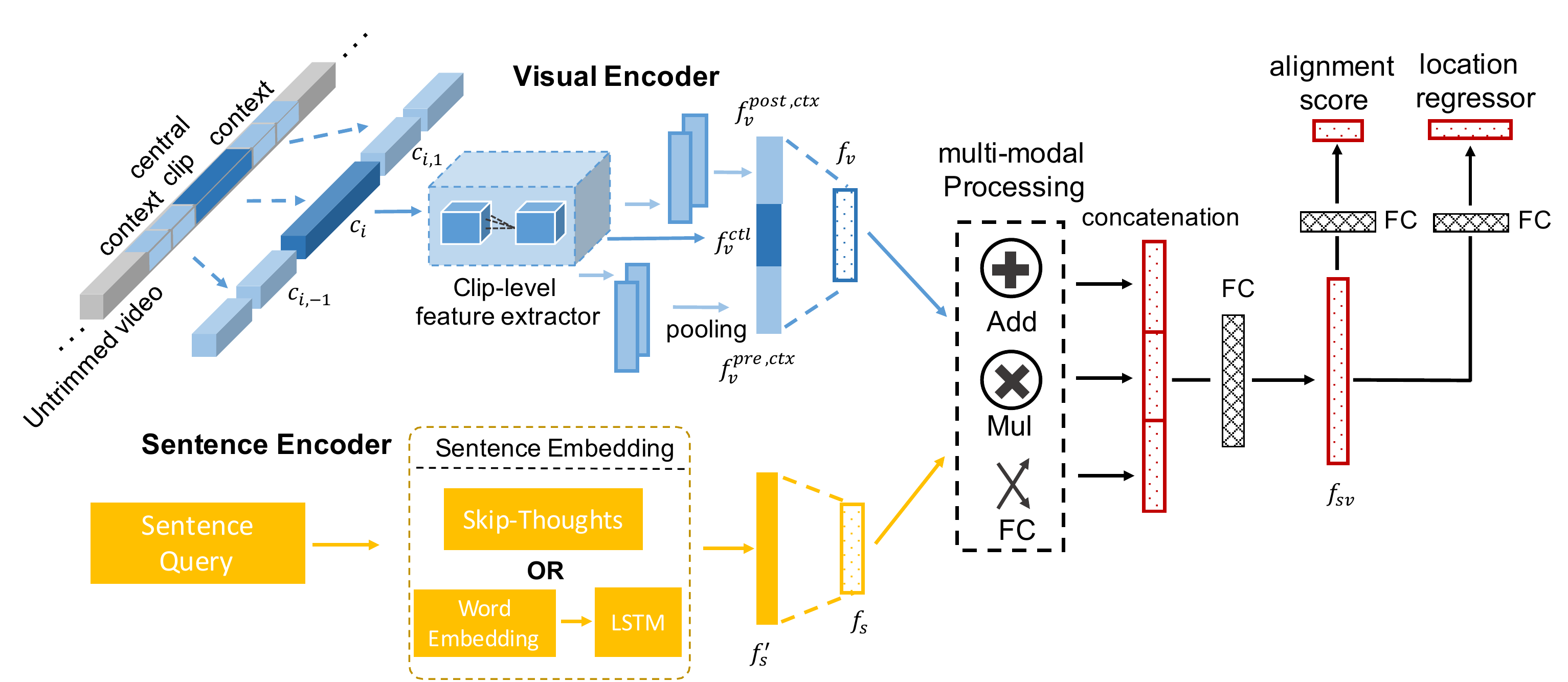}
\caption{ Cross-modal Temporal Regression Localizer (CTRL) architecture.  CTRL contains four modules: a visual encoder to extract features for video clips, a sentence encoder to extract embeddings, a multi-modal processing network to generate combined representations for visual and text domain, and a temporal regression network to produce alignment scores and location offsets.}
\end{figure*}

In this section, we describe our Cross-modal Temporal  Regression Localizer (CTRL) for Temporal Activity Localization via Language (TALL) and training procedure in detail.  CTRL contains four modules: a visual encoder to extract features for video clips, a sentence encoder to extract embeddings, a multi-modal processing network to generate combined representations for visual and text domain, and a temporal regression network to produce alignment scores and location offsets between the input sentence query and video clips.

\subsection{Problem Formulation}
We denote a video as $V=\{f_t\}_{t=1}^{T}$, $T$ is the frame number of the video. Each video is associated with temporal sentence annotations: $A=\{(s_j,\tau_j^s,\tau_j^e)\}_{j=1}^{M}$, M is the sentence annotation number of the video $V$, $s_j$ is a natural language sentence of a video clip, which has $\tau_j^s$ and $\tau_j^e$ as start and end time in the video. The training data are the sentence and video clip pairs. The task is to predict one or more $(\tau_j^s, \tau_j^e)$ for the input natural language sentence query.

\subsection{CTRL Architecture}

\textbf{Visual Encoder.}
For a long untrimmed video $V$,  we generate a set of video clips $C=\{(c_i, t_i^s, t_i^e)\}_{i=1}^H$ by temporal sliding windows, where $H$ is the total number of the clips of the video $V$, $t_i^s$ and $t_i^e$ are the start and end time of video clip $c_i$.  We define visual encoder as a function $F_{ve}(c_i)$ that maps a certain clip $c_i$ and its context to a feature vector $f_v$, whose dimension is $d_s$. Inside the visual encoder, a feature extractor $E_{v}$ is used to extract clip-level feature vectors, whose input is $n_f$ frames and output is a vector with dimension $d_v$. 
For one video clip $c_i$, we consider itself (as the central clip) and its surrounding clips (as context clips) $ c_{i,q}, q \in [-n,n]  $, $j$ is the clip shift, $n$ is the shift boundary. We uniformly sample $n_f$ frames from each clip (central and context clips). The feature vector of central clip is denoted as $f_{v}^{ctl}$. For the context clips, we use a pooling layer to calculate a pre-context feature $f_{v}^{pre}=\frac{1}{n}\sum_{q=-n}^{-1}E_{v}(c_{i,q})$ and post-context feature $f_{v}^{post}=\frac{1}{n}\sum_{q=1}^{n}E_{v}(c_{i,q})$. 
Pre-context feature and post-context feature are pooled separately, as the end and the start of an activity can be quite different and both could be critical for temporal localization. $f_{v}^{pre}$, $f_{v}^{ctl}$ and  $f_{v}^{post}$ are concatenated and then linearly transformed to the feature vector $f_{v}$ with dimension $d_s$, as the visual representation for clip $c_i$.

\textbf{Sentence Encoder.}
A sentence encoder is a function $F_{se}(s_j)$ that maps a sentence description $s_j$ to a embedding space, whose dimension is $d_s$( the same as visual feature space ).  Specifically, a sentence embedding extractor $E_s$ is used to extract a sentence-level embedding $f'_s$ and is followed by a linear transformation layer, which maps $f'_s$ to $f_s$ with dimension $d_s$, the same as visual representation $f_v$.
We experiment two kinds of sentence embedding extractors, one is a LSTM network which takes a word as input at each step, and the hidden state of final step is used as sentence-level embedding; the other is an off-the-shelf sentence encoder, Skip-thought \cite{kiros2015skip}. More details would be discussed in Section 4. 

\textbf{Multi-modal Processing Module.}
The inputs of the multi-modal processing module are a visual representation  $f_v$ and a sentence embedding $f_s$, which have the same dimension $d_s$. We use vector element-wise addition ($+$), vector element-wise multiplication ($\times$) and vector concatenation ($\parallel$) followed by a Fully Connected ($FC$) layer to combine the information from both modalities. Addition and multiplication operation allow additive and multiplicative interaction between two modalities and don't change the feature dimension. The $FC$ layer allows interaction among all elements. The input dimension of the $FC$ layer is $2*d_s$ and the output dimension is $d_s$. The outputs from all three operations are concatenated to construct a multi-modal representation $f_{sv}=(f_{s}\times f_{v}) \parallel (f_{s}+f_{v}) \parallel FC(f_s\parallel f_v)$, which is the input for our core module, temporal localization regression networks.

\textbf{Temporal Localization Regression Networks.}
Temporal localization regression network takes the multi-modal representation $f_{sv}$ as input, and has two sibling output layers. The first one outputs an alignment score $cs_{i,j}$ between the sentence $s_j$ and the video clip $c_i$. The second one outputs clip location regression offsets. We design two location offsets, the first one is parameterized offset:  $t=(t_c, t_l)$, where $t_c$ and $t_l$ are parameterized central point offset and length offset respectively. The parameterization is as follows:
\begin{equation}
t_p=(p-p_c)/l_c,  \ t_l=log(l/l_c)
\end{equation}
where $p$ and $l$ denote the clip's center coordinate and clip length respectively. Variables $p, p_c$ are for predicted clip and test clip (like wise for $l$). The second offset is non-parameterized offset: $t=(t_s,t_e)$, where $t_s$ and $t_e$ are the start and end point offsets. 
\begin{equation}
t_s=s-s_c, \  t_e=e-e_c
\end{equation}
where $s$ and $e$ denote the clip's start and end coordinate respectively. Temporal coordinate regression can be thought as clip location regression from a test clip to a nearby ground-truth clip, as the original clip could be either too tight or too loose, the regression process tend to find better locations. 


\subsection{CTRL Training}
\textbf{Multi-task Loss Function.}
CTRL contains two sibling output layers, one for alignment and the other for regression. We design a multi-task loss $L$ on a mini-batch of training samples to jointly train for visual-semantic alignment and clip location regression.
\begin{equation}
L=L_{aln}+\alpha L_{reg}
\end{equation}
where $L_{aln}$ is for visual-semantic alignment and $L_{reg}$ is for clip location regression, and $\alpha$ is a hyper-parameter, which controls the balance between the two task losses. The alignment loss encourages aligned clip-sentence pairs to have positive scores and misaligned pairs to have negative scores.
\begin{align}
L_{aln}=\frac{1}{N}\sum_{i=0}^{N} [\alpha_{c}\text{log}(1+\text{exp}(-cs_{i,i}))+ \notag\\
 \sum_{j=0,j\neq i}^{N}\alpha_{w}\text{log}(1+\text{exp}(cs_{i,j}))]
\end{align}
where $N$ is the batch size, $cs_{i,j}$ is the alignment score between sentence $s_j$ and video clip $c_i$, $\alpha_c $ and $\alpha_w$ are the hyper parameters which control the weights between positive ( aligned ) and negative ( misaligned ) clip-sentence pairs. 

The regression loss $L_{reg}$ is calculated for the aligned clip-sentence pairs. A sentence $s_j$ annotation contains start and end time $(\tau_j^s, \tau_j^e)$. The aligned sliding window clip $c_i$ has  $(t_i^s, t_i^e)$.  The ground truth offsets $t^*$ are calculated from start and end times.
\begin{equation}
L_{reg}=\frac{1}{N}\sum_{i=0}^{N}[R(t_{x,i}^*-t_{x,i})+R(t_{y,i}^*-t_{y,i})]
\end{equation}
where $x$ and $y$ indicate $p$ and $l$ for parameterized offsets, or $s$ and $e$ for non-parameterized offsets. $R(t)$ is smooth $L_1$ function.

\textbf{Sampling Training Examples.}
To collect training samples, we use multi-scale temporal sliding windows with [64, 128, 256, 512] frames and 80\% overlap.  (Note that, at test time,  we only use coarsely sampled clips.) 
We use the following strategy to collect training samples $T=\{[(s_h,\tau_h^s,\tau_h^e),(c_h, t_h^s, t_h^e)]\}_{h=0}^{N_T}$. Each training sample contains a sentence description $(s_h,\tau_h^s,\tau_h^e)$ and a video clip $(c_h, t_h^s, t_h^e)$. For a sliding window clip $c$ from $C$ with temporal annotation $(t^s, t^e)$ and a sentence description $s$ with temporal annotation $(\tau^s,\tau^e)$,  we align them as a pair of training samples if they satisfy (1) Intersection over Union (IoU) is larger than 0.5; (2) non Intersection over Length (nIoL) is smaller than 0.2 and (3) one sliding window clip can be aligned with only one sentence description. The reason we use nIoL is that  we want the the most part of the sliding window clip to overlap with the assigned sentence, and simply increasing IoU threshold would harm regression layers ( regression aims to move the clip from low IoU to high IoU). As shown in Figure 3, although the IoU between $c$ and $s_1$ is about 0.5, if we assign $c$ to $s_1$, then it will disturb the model ,because $c$ contains information of $s_2$. 
\begin{figure}
\centering
\includegraphics[scale=0.65]{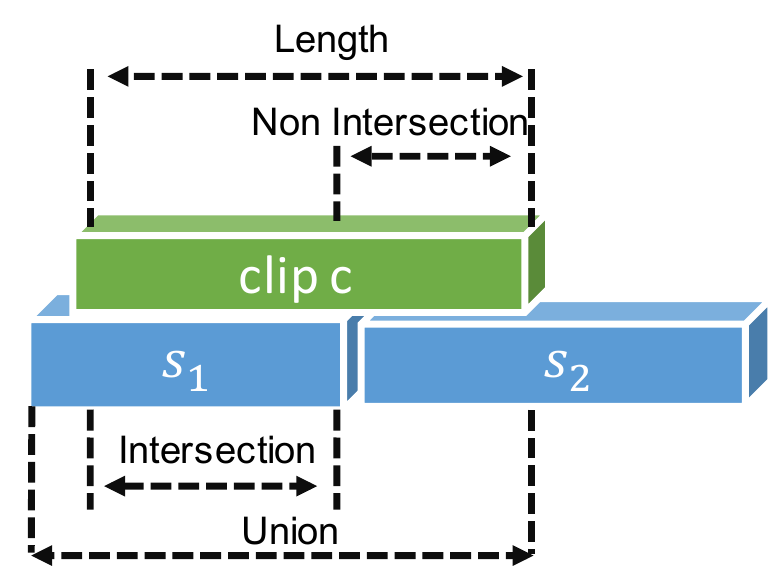}
\caption{ Intersection over Union (IoU) and non-Intersection over Length (nIoL). }
\end{figure}


\section{Evaluation}
In this section, we describe the evaluation settings and discuss the experiment results 
\subsection{Datasets}
\textbf{TACoS \cite{regneri2013grounding}.} This dataset was built on the top of MPII-Compositive dataset \cite{rohrbach2012script} and contains 127 videos. Every video is associated with two type of annotations. The first one is fine-grained activity labels with temporal location (start and end time). The second set of annotations is natural language descriptions with temporal locations. The natural language descriptions were obtained by crowd-sourcing annotators, who were asked to describe the content of the video clips by sentences. In total, there are 17344 pairs of sentence and video clips. We split it in 50\% for training, 25\% for validation and 25\% for test.

\textbf{Charades-STA.} Charades \cite{sigurdsson2016hollywood} contains around 10k videos and each video contains temporal activity annotation (from 157 activity categories) and multiple video-level descriptions. TALL needs clip-level sentence annotation: sentence descriptions with start and end time, which are not provided in the original Charades dataset. We noticed that the names of activity categories in Charades are parsed from the video-level descriptions, so many of activity names appear in descriptions. Another observation we make is that most descriptions in Charades share a similar syntactic structure: consisting of multiple sub-sentences, which are connected by comma, period and conjunctions, such as ``then", ``while", ``after", ``and".  For example, ``A person is sitting down by the door. They stand up and start carefully leaving some dishes in the sink". 

Based on these observations, we designed a semi-automatic way to generate sentence temporal annotation. The first step is sentence decomposition: a long sentence is split to sub-sentences by a set of conjunctions (which are collected by hand ), and for each sub-sentence, the subject ( parsed by Stanford CoreNLP \cite{manning2014stanford} ) of the original long sentence is added to start. The second step is keyword matching: we extract keywords for each activity categories and match them to sub-sentences, if they are matched, the temporal annotation (start and end time) are assigned to the sub-sentences. The third step is a human check: for each pair of sub-sentence and temporal annotation, we (two of the co-authors) checked whether the sentence made sense and whether they matched the activity annotation. An example is shown in Figure 4. 

Although TACoS and Charades-STA are challenging, their lengths of queries are limited to single sentences. To explore the potential of CTRL framework on handling longer and more complex sentences,  we build a complex sentence set. Inside each video, we connect consecutive sub-sentences to make complex query, each complex query contains at least two sub-sentences, and is checked to make sure that the time span is less than half of the video length. We use them for test purpose only. In total, there are 13898 clip-sentence pairs in Charades-STA training set, 4233 clip-sentence pairs in test set  and 1378 complex sentence quires. On average, there are 6.3 words per non-complex sentence, and 12.4 words per complex sentence.

\begin{figure}
\centering
\includegraphics[scale=0.40]{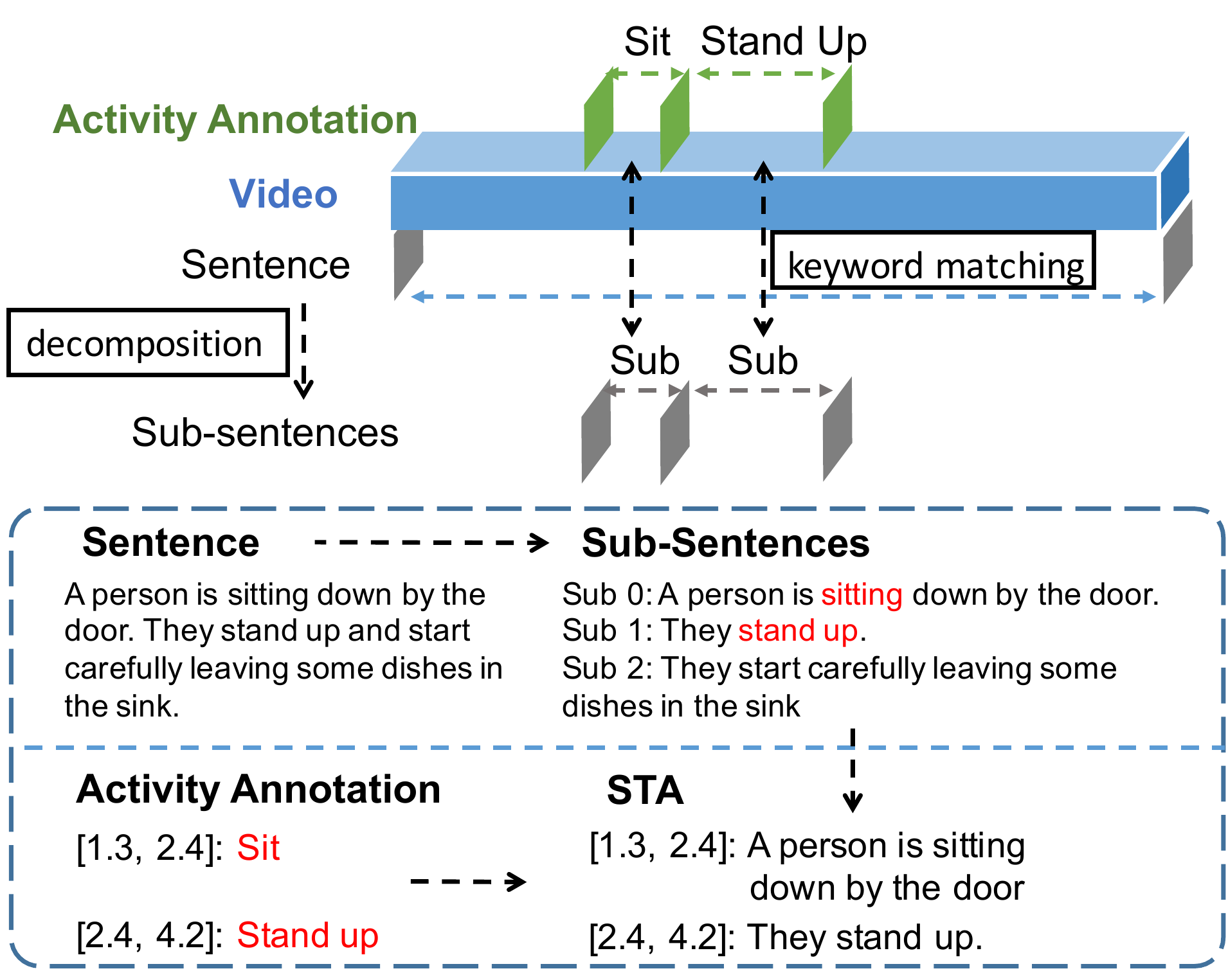}
\caption{ Charades-STA construction. }
\end{figure}


\subsection{Experiment Settings}
We will introduce evaluation metric, baseline methods and our system variants in this part.
\subsubsection{Evaluation Metric}
We adopted a similar metric used by \cite{hu2015natural} to compute ``R@$n$, IoU=$m$", which means that the percentage of at least one of the top-$n$ results having Intersection over Union (IoU) larger than $m$. This metric itself is on sentence level, so the overall performance is the average among all the sentences. $R(n, m)=\frac{1}{N} \sum_{i=1}^{N} r(n, m, s_i)$, 
where $r(n,m,s_i)$ is the recall for a query $s_i$, $N$ is total number of queries and $R(n,m)$ is the averaged overall performance.

\subsubsection{Baseline Methods}
We consider two sentence based image/video retrieval baseline methods: visual-semantic alignment with LSTM (VSA-RNN ) \cite{karpathy2015deep} and visual-semantic alignment with Skip-thought vector (VSA-STV) \cite{kiros2015skip}. For these two baseline methods, we use the same training samples and test sliding windows as those for CTRL. 

\textbf{VSA-RNN.} This baseline method is similar to the model in DVSA \cite{karpathy2015deep}. We use a regular LSTM instead of BRNN to encode the input description. The size of hidden state of LSTM is 1024 and the output size is 1000. Video clips are processed by a C3D network that is pre-trained on Sports1M \cite{Karpathy_2014_CVPR}. The 4096 dimensional $fc6$ vector is extracted and linearly transformed to 1000 dimensional, which is used as the clip-level feature. Cosine similarity is used to calculate the confidence score between the clip and the sentence. Hinge loss is used to train the model. At test time, we compute the alignment score between input sentence query and all the sliding windows in the video. \textbf{VSA-STV:} Instead of using RNN to extract sentence embedding, we use an off-the-shelf Skip-thought \cite{kiros2015skip} sentence embedding extractor. A skip-thought vector is 4800 dimensional, we linearly transform it to 1000 dimensional. Visual encoder is the same as for VSA-RNN.

\textbf{Verb and Object Classifiers.} We also implemented baseline methods based on annotations of pre-defined actions and objects. TACoS dataset also contains pre-defined actions and object annotations at clip-level. These objects and actions annotations are from the original MPII-Compositive dataset \cite{rohrbach2012script}. 54 categories of actions and 81 categories of objects are involved in training set. We use the same C3D feature as above to train action classifiers and object classifiers. The classifier is based on a 2-layer fully connected network, the size of first layer is 4094 and the size of second layer is the number of categories. The test sentences are parsed by Stanford CoreNLP \cite{manning2014stanford}, and verb-object (VO) pairs are extracted using the sentence dependencies. The VO pairs are matched with action and object annotations based on string matching. The alignment score between a sentence query and a clip is the score of matched action and object classifier responses. \textbf{Verb} means that we only use action classifier; \textbf{Verb+Obj} means that both action classifiers and object classifiers are used.

\subsubsection{System Variants}
We experimented with variants of our system to test the effectiveness of our method. \textbf{CTRL(aln)}: we don't use regression, train the CTRL with only alignment loss $L_{aln}$. \textbf{CTRL(reg-p)}: train the CTRL with alignment loss $L_{aln}$ and parameterized regression loss $L_{reg-p}$. \textbf{CTRL(reg-np):} context information is considered and CTRL is trained with alignment loss $L_{aln}$ and non-parameterized regression loss $L_{reg-np}$. \textbf{CTRL(loc):} SCNN \cite{Shou_2016_CVPR} proposed to use overlap loss to improve activity localization performance. Based on our pure alignment(without regression), we implemented a similar loss function considering clip overlap as in SCNN. $L_{loc}=\sum_i^n (0.5*( \frac{1/(1+e^{-cs_{i,i}})^2}{IoU_i}-1 ))$, 
where $cs_{i,i}$ and $IoU_i$ are respectively the alignment score and Intersection over Union (IoU) between the aligned pairs of sentence and clip in a mini-batch. The major difference is that SCNN solved a classification problem, so they use Softmax score, however in our case, we consider an alignment problem. The overall loss function is $L_{scnn}=L_{aln}+L_{loc}$. For this method, we use C3D as the visual encoder and Skip-thought as the sentence encoder.

\subsection{Experiments on TACoS}
In this part, we discuss the experiment results on TACoS. First we compare the performance of different visual encoders; second we compare two sentence embedding methods; third we compare the performance of CTRL variants and baseline methods. The length of sliding windows for test is 128 with overlap 0.8, multi-scale windows are not used. We empirically set the context clip number $n$ as $1$ and the length of context window as $128$ frames. The dimension of $f_v$, $f_s$ and $f_{sv}$ are all set to 1000. We set batch size as 64, the networks are optimized by Adam \cite{kingma2014adam} optimizer on a Nvidia TITAN X GPU.

\begin{figure}
\centering
\includegraphics[scale=0.38]{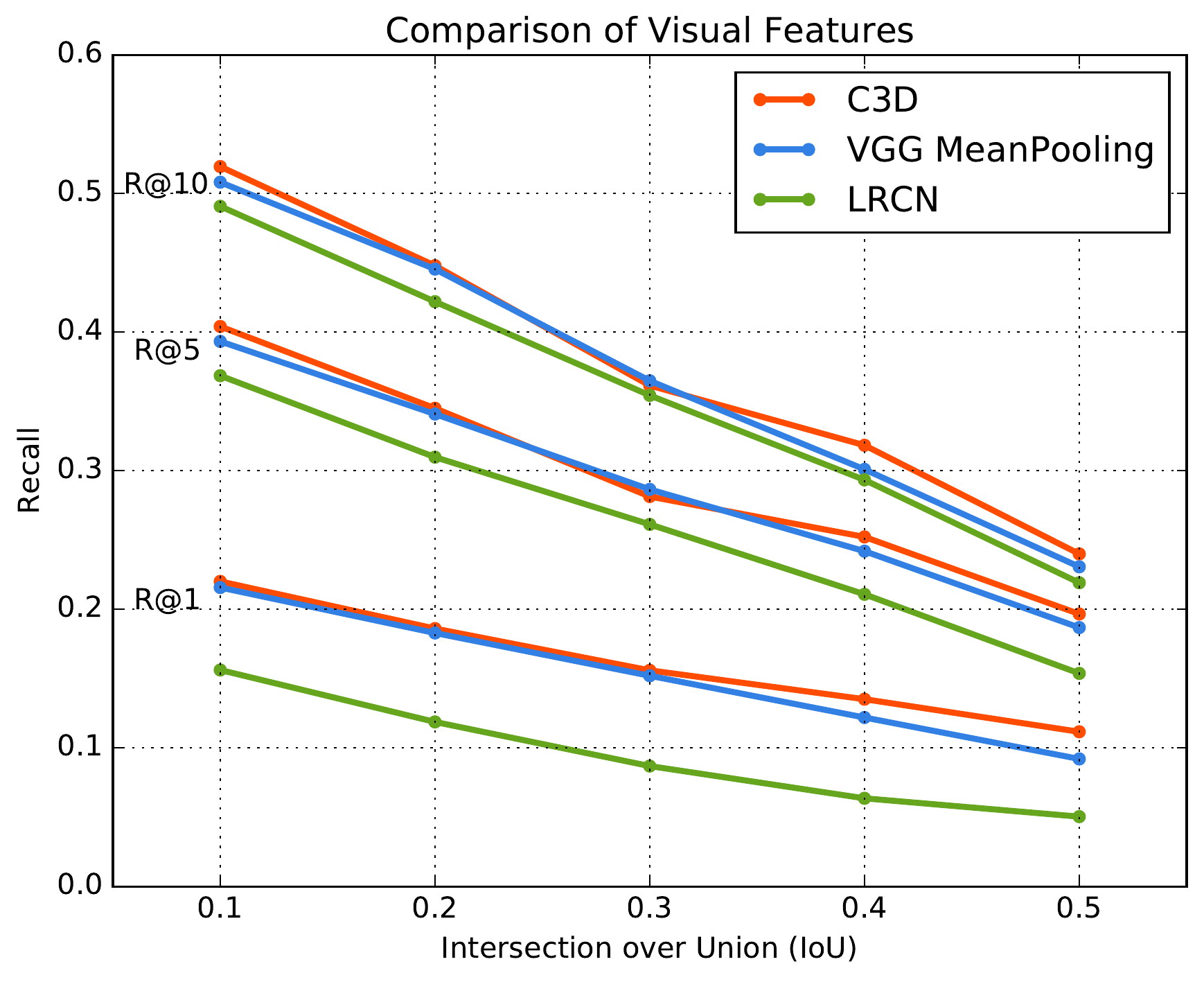}
\caption{ Performance comparison of different visual encoders. }
\end{figure}

\textbf{Comparison of visual features.} We consider three clip-level visual encoders: C3D \cite{tran2015learning}, LRCN \cite{donahue2015long}, VGG+Mean Pooling \cite{Karpathy_2014_CVPR}. Each of them takes a clip with 16 frames as input and outputs a 1000-dimensional feature vector. For C3D, $fc6$ feature vector is extracted and then linearly transformed to 1000-dimension.  For LRCN and VGG poolng, we extract $fc6$ of  VGG-16 for each frame. The LSTM's hidden state size is 256.
We use Skip-thought as the sentence embedding extractor and other parts of the model are the same to CTRL(aln). There are three groups of curves, which are Recall@10, Recall@5 and Recall@1 respectively, shown in Figure. 5.  We can see that C3D performs generally better than other two methods. LRCN's performance is inferior, the reason maybe that the dataset is relatively small, not enough to train the LSTM well.

\textbf{Comparison of sentence embedding.} For sentence encoder, we consider two commonly used methods: word2vec+LSTM \cite{hu2015natural} and Skip-thought \cite{kiros2015skip}. In our implementation of word2vec, we train skip-gram model on English Dump of Wikipedia. The dimension of the word vector is 500 and the hidden state size of the LSTM is 512. For Skip-thought vector, we linearly transform it from 4800-dimension to 1000-dimension. We use C3D as the  visual feature extractor and other parts are the same to CTRL(aln).
\begin{figure}
\centering
\includegraphics[scale=0.43]{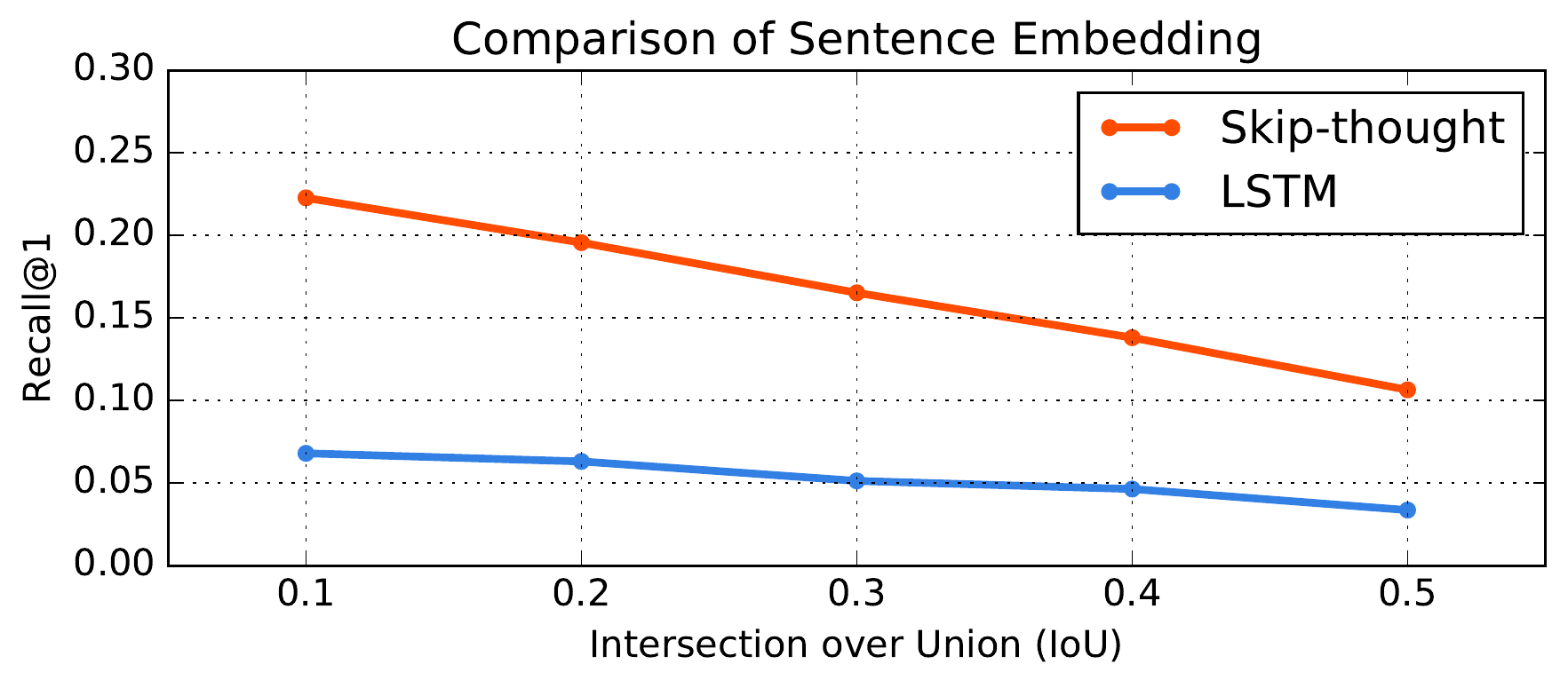}
\caption{ Performance comparison of different sentence embedding. }
\end{figure}
From the results, we can see that the performance of Skip-thought is generally better than word2vec+LSTM. We conjecture the reason is that the scale of TACoS is not large enough to train the LSTM (comparing with the counterpart datasets in object detection, like ReferIt \cite{kazemzadeh2014referitgame}, Flickr30k Entities \cite{plummer2015flickr30k}, which contains over 100k sentences).

\textbf{Comparison with other methods.} We test our system variants and baseline methods on TACoS and report the result for $IoU\in\{0.1,0.3,0.5\}$ and Recall@\{1, 5\}. The results are shown in Table 1. ``Random" means that we randomly select $n$ windows from the test sliding windows and evaluate Recall@$n$ with IoU=$m$. All methods use the same C3D features. VSA-RNN uses the end-to-end trained LSTM as the sentence encoder and all other methods use pre-trained Skip-thought as sentence embedding extractor. 

We can see that visual retrieval baselines (\emph{i.e.} VSA-RNN, VSA-STV) lead to inferior performance, even compared with our pure alignment model CTRL(aln). We believe the major reasons are two-fold: 1) the multilayer alignment network learns better alignment than the simple cosine similarity model, which is trained by hinge loss function; 2) visual retrieval models do not encode temporal context information in a video.  Pre-defined classifiers also produce inferior results. We think it is mainly because the pre-defined actions and objects are not precise enough to represent sentence queries. By comparing Verb and Verb+Obj, we can see that additional object (such as ``knife", ``egg") information helps to represent sentence queries. 

\textbf{Temporal action boundary regression} As described before, we implemented a temporal localization loss function similar to the one in SCNN \cite{Shou_2016_CVPR}, which consider clip overlap. Experiment results show that CTRL(loc) does not bring much improvement over CTRL(aln), perhaps because CTRL(loc) still relies on clip selection from sliding windows, which may not overlap with ground truth well. CTRL(reg-np) outperforms CTRL(aln) and CTRL(loc) significantly, showing the effectiveness of temporal regression model.  By comparing CTRL(reg-p) and CTRL(reg-np) in Table 1, it can be seen that non-parameterized setting helps the localizer regress the action boundary to a more accurate location. We think the reason is that unlike objects can be re-scaled in images due to camera projection, actions' time spans can not be easily rescaled in videos (we don't consider slow motion and quick motion). Thus, to do the boundary regression effectively, the object bounding box coordinates should be first normalized to some standard scale,  but for actions, time itself is the standard scale.

Some prediction and regression results are shown in Figure 7. We can see that the alignment prediction gives a coarse location, which is limited by the fixed window length; the regression model helps to refine the clip's bounding box to a higher IoU location. 

\setlength{\tabcolsep}{2.5pt}
\begin{table}\footnotesize
\centering
\caption{Comparison of different methods on TACoS}
\begin{tabular}{| l| c c c c c c |}

\hline
Method &\multicolumn{1}{c}{\begin{tabular}[c]{@{}c@{}}R@1\\ IoU=0.5\end{tabular}}&\multicolumn{1}{c}{\begin{tabular}[c]{@{}c@{}}R@1\\ IoU=0.3\end{tabular}}&\multicolumn{1}{c}{\begin{tabular}[c]{@{}c@{}}R@1\\ IoU=0.1\end{tabular}}&\multicolumn{1}{c}{\begin{tabular}[c]{@{}c@{}}R@5\\ IoU=0.5\end{tabular}}&\begin{tabular}[c]{@{}c@{}}R@5\\ IoU=0.3\end{tabular}&\begin{tabular}[c]{@{}c@{}}R@5\\ IoU=0.1\end{tabular}\\ \hline
Random& 0.83&1.81&3.28 &3.57&7.03&15.09 \\ \hline
Verb& 1.62&2.62&6.71 &3.72&6.36&11.87 \\
Verb+Obj& 8.25&11.24&14.69 &16.46&21.50&26.60 \\ \hline
VSA-RNN& 4.78&6.91&8.84 &9.10&13.90&19.05 \\
VSA-STV & 7.56&10.77&15.01&15.50&23.92&32.82\\  \hline
CTRL (aln) & 10.67&16.53&22.29&19.44&29.09&41.05\\
CTRL (loc) &10.70&16.12&22.77&18.83&31.20&45.11\\
CTRL (reg-p) & 11.85&17.59&23.71&23.05&33.19&47.51\\
CTRL (reg-np) & \textbf{13.30}&\textbf{18.32}&\textbf{24.32}&\textbf{25.42}&\textbf{36.69}&\textbf{48.73}\\
\hline\end{tabular}

\end{table}

\subsection{Experiments on Charades-STA}
In this part, we evaluate CTRL models and baseline methods on Charades-STA and report the results for $IoU\in\{0.5, 0.7\}$ and Recall@\{1, 5\}, which are shown in Table 2. The lengths of sliding windows for test are 128 and 256, window's overlap is 0.8. It can be seen that the results are consistent with those in TACoS. CTRL(reg-np) shows a significant improvement over CTRL(aln) and CTRL(loc). The non-parameterized settings (CTRL(reg-np)) work consistently better than the parameterized settings (CTRL(reg-p)). Figure 8 shows some prediction and regression results. 

\setlength{\tabcolsep}{4pt}
\begin{table}\small
\centering
\caption{Comparison of different methods on Charades-STA}
\begin{tabular}{| l| c c c c |}

\hline
Method &\multicolumn{1}{c}{\begin{tabular}[c]{@{}c@{}}R@1\\ IoU=0.5\end{tabular}}&\multicolumn{1}{c}{\begin{tabular}[c]{@{}c@{}}R@1\\ IoU=0.7\end{tabular}}&\multicolumn{1}{c}{\begin{tabular}[c]{@{}c@{}}R@5\\ IoU=0.5\end{tabular}}&\begin{tabular}[c]{@{}c@{}}R@5\\ IoU=0.7\end{tabular}\\ \hline
Random& 8.51 &3.03&37.12&14.06 \\ \hline
VSA-RNN& 10.50&4.32&48.43 &20.21 \\
VSA-STV & 16.91&5.81&53.89&23.58\\  \hline
CTRL (aln) & 18.77&6.53&54.29&23.74\\
CTRL (loc) &20.19&6.92&55.72&24.41\\
CTRL (reg-p) & 22.27&8.46&57.83&26.61\\
CTRL (reg-np) & \textbf{23.63}&\textbf{8.89}&\textbf{58.92}&\textbf{29.52}\\
\hline\end{tabular}

\end{table}

\begin{table}\small
\centering
\caption{Experiments of complex sentence query.}
\begin{tabular}{| l| c c c c |}

\hline
Method &\multicolumn{1}{c}{\begin{tabular}[c]{@{}c@{}}R@1\\ IoU=0.5\end{tabular}}&\multicolumn{1}{c}{\begin{tabular}[c]{@{}c@{}}R@1\\ IoU=0.7\end{tabular}}&\multicolumn{1}{c}{\begin{tabular}[c]{@{}c@{}}R@5\\ IoU=0.5\end{tabular}}&\begin{tabular}[c]{@{}c@{}}R@5\\ IoU=0.7\end{tabular}\\ \hline
Random& 11.83 &3.21&43.28&18.17 \\ \hline
CTRL& 24.09&8.03&69.89 &32.28 \\
CTRL+Fusion & 25.82&8.32&69.94&32.81\\  \hline

\hline\end{tabular}
\end{table}

\begin{figure*}
\centering
\includegraphics[scale=0.40]{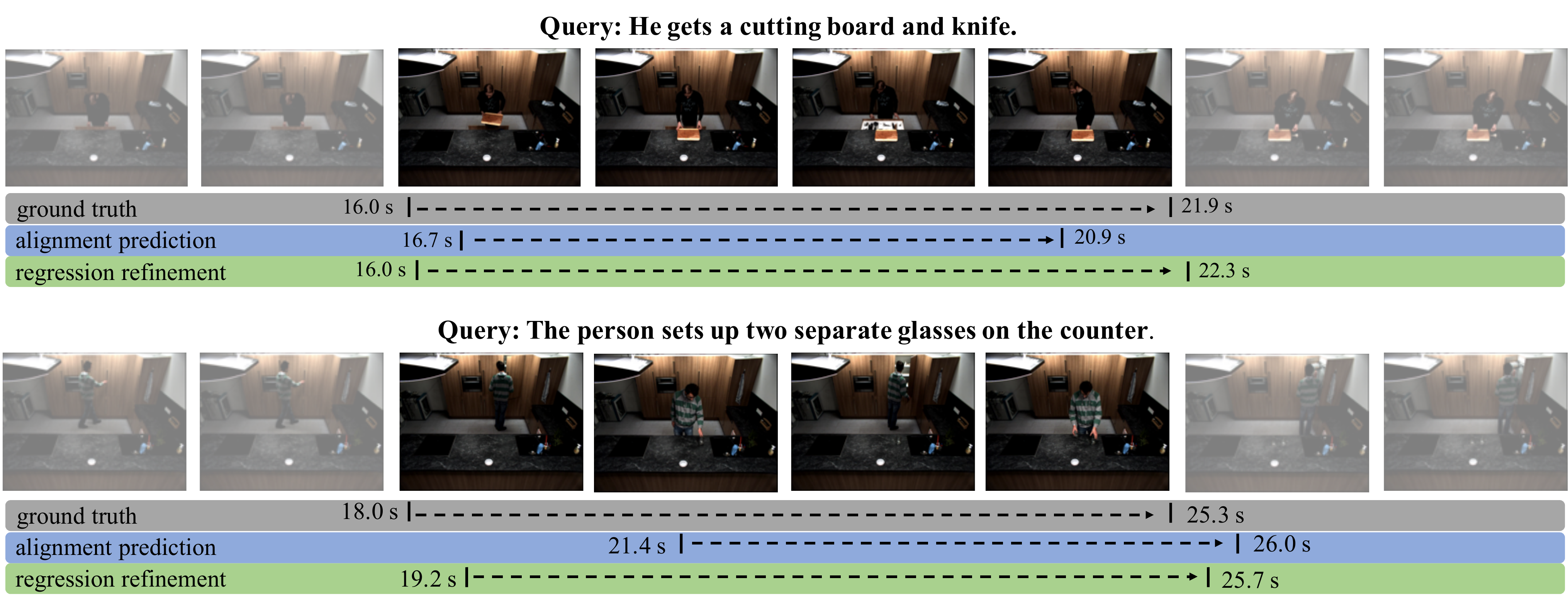}
\caption{ Alignment prediction and regression refinement examples in TACoS. The row with gray background shows the ground truth for the given query; the row with blue background shows the sliding window alignment results; the row with green background shows the clip regression results.}
\end{figure*}
\begin{figure*}
\centering
\includegraphics[scale=0.41]{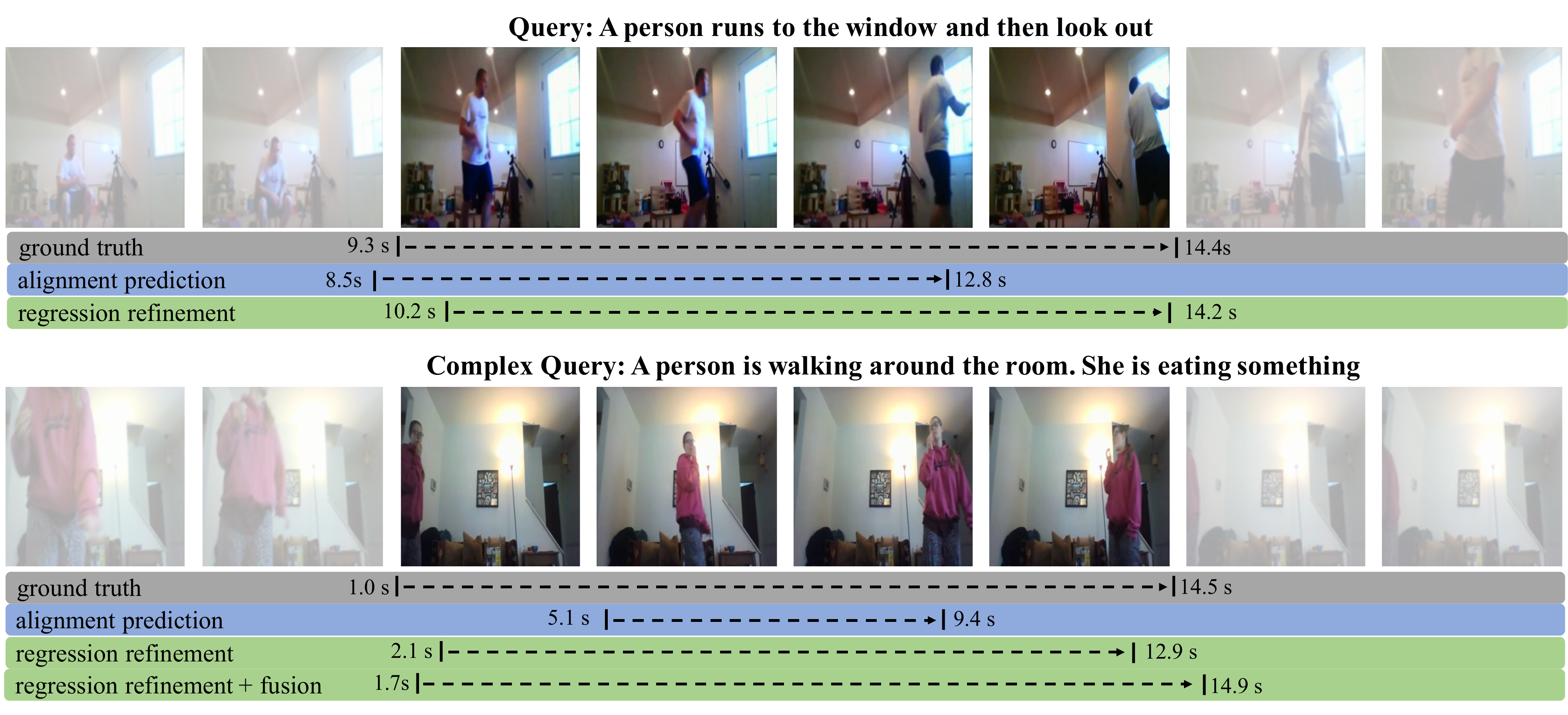}
\caption{ Alignment prediction and regression refinement examples in Charades-STA.}
\end{figure*}

We also test complex sentence query on Charades-STA. As shown in Table. 3, ``CTRL" means that we simply input the whole complex sentence into CTRL model. ``CTRL+fusion" means that we input each sentence of a complex query separately into CTRL, and then do a late fusion. Specifically, we compute the average alignment score over all sentences, take the minimum of all start times and maximum of all end times as start and end time of the complex query. 
Although the random performance in Table. 3 (complex) is higher than that in Table 2 (single), the gain over random performance remains similar, which indicates that CTRL is able to handle complex query consisting multiple activities well. Comparing CTRL and CTRL+Fusion, we can see that CTRL could be an effective first step for complex query, if combined with other fusion methods.

In general, we observed two types of common hard cases: (1) long query sentences increase chances of failure, likely because the sentence embeddings are not discriminative enough;  (2) videos that contain similar activities but with different objects (e.g. in TACOS dataset, put a cucumber on chopping board, and put a knife on chopping board) are hard to distinguish amongst each other.

\section{Conclusion}
We addressed the problem of Temporal Activity Localization via Language (TALL) and proposed a novel Cross-modal Temporal Regression Localizer (CTRL) model, which uses temporal regression for activity location refinement. We showed that non-parameterized offsets works better than parameterized offsets for temporal boundary regression. 
Experimental results show the effectiveness of our method on TACoS and Charades-STA. 

{\small
\bibliographystyle{ieee}
\bibliography{egbib}
}

\clearpage
\section{Supplementary Material}
To directly compare our the temporal regression method with previous state-of-the-art methods on traditional action detection task, we did additional experiments on THUMOS-14.

Since THUMOS is a classification task with limited number of action classes, we removed the cross-modal part and trained the localization network with classification loss (cross-entropy loss) and regression loss. We trained a model on the validation set (train set only contains trimmed videos which are not suitable for localization task) and tested it on the test set. The regression model contains 20*2 outputs, corresponding to the 20 categories in the dataset, $\alpha$ is set to 2.0 and 10.0 for non-parameterized and parameterized regression respectively. For each category, we use NMS to eliminate redundant detections in every video, the NMS threshold is set to (tIoU - delta), where tIoU = 0.5 and delta=0.2. We report mAP at tIoU=0.5. For training sample generation, we use the same procedure as SCNN [24], we set the high IoU threshold as 0.5 (SCNN used 0.7) and low IoU threshold as 0.1 (SCNN used 0.3) for generating training samples. Note that, our method and SCNN both use C3D features.

\begin{table}[h]
\centering
\caption{Temporal action localization experiments on THUMOS-14}
\label{my-label}
\begin{tabular}{l|lllll}
\hline
    & SCNN & cls  & reg-p & reg-np & reg-np (p+d) \\ \hline
mAP & 19.0 & 16.3 & 18.9  & 19.8   & 20.5                        \\ \hline
\end{tabular}
\end{table}

As shown, ``cls'' for only using classification loss, ``reg-p" for classification loss+parameterized regression loss, ``reg-np" for classification loss+ non-parameterized regression loss. For ``cls'', ``reg-p'',``reg-np'', we use the proposals generated by SCNN (from their github codes) as input, so that we can fairly compare the effect of classification loss, localization loss (used in SCNN) and temporal regression loss. ``reg-np (p+d)" means that we apply temporal regression on both proposal generation and action detection.

Our method (reg-np) outperforms SCNN. Comparing with ``cls" and ``reg-np", we can see the improvement by the temporal regression. By applying temporal regression on proposal generation, we can see a further improvement from 19.8 to 20.5.

\end{document}